\documentclass{article}

\usepackage{arxiv}

\usepackage[utf8]{inputenc} 
\usepackage[T1]{fontenc}    
\usepackage{hyperref}       
\usepackage{url}            
\usepackage{booktabs}       
\usepackage{amsfonts}       
\usepackage{nicefrac}       
\usepackage{microtype}      
\usepackage{lipsum}		
\usepackage{graphicx}
\usepackage{natbib}
\usepackage{doi}
\usepackage{tabulary}
\usepackage{enumitem}

\setcitestyle{numbers}

\title{Machine Learning based Anomaly Detection for Smart Shirt: A Systematic Review}

\date{March 4, 2022}	

\author{ \href{https://orcid.org/0000-0002-5345-8854}{\includegraphics[scale=0.06]{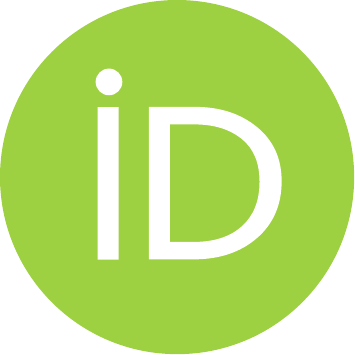}\hspace{1mm}Eduardo C. Nunes} \\
	Tech Team\\
	Mountains of Research Collaborative Laboratory\\
	Braganca, Portugal \\
	\texttt{enunes@morecolab.pt} \\
}

\renewcommand{\shorttitle}{\textit{arXiv} Template}

\hypersetup{
pdftitle={Machine Learning based Anomaly Detection for Smart Shirt},
pdfsubject={q-bio.NC, q-bio.QM},
pdfauthor={David S.~Hippocampus, Elias D.~Striatum},
pdfkeywords={First keyword, Second keyword, More},
}

\begin{document}
\maketitle

\begin{abstract}
In recent years, the popularity and use of Artificial Intelligence (AI) and large investments on the Internet of Medical Things (IoMT) will be common to use products such as smart socks, smart pants, and smart shirts. These products are known as Smart Textile or E-textile, which has the ability to monitor and collect signals that our body emits. These signals make it possible to extract anomalous components using Machine Learning (ML) techniques that play an essential role in this area. This study presents a Systematic Review of the Literature (SLR) on Anomaly Detection using ML techniques in Smart Shirt. The objectives of the SLR are: (i) to identify what type of anomaly the smart shirt; (ii) what ML techniques are being used; (iii) which datasets are being used; (iv) identify smart shirt or signal acquisition devices; (v) list the performance metrics used to evaluate the ML model; (vi) the results of the techniques in general; (vii) types of ML algorithms are being applied.The SLR selected 11 primary studies published between 2017-2021. The results showed that 6 types of anomalies were identified, with the Fall anomaly being the most cited. The Support Vector Machines (SVM) algorithm is most used. Most of the primary studies used public or private datasets. The Hexoskin smart shirt was most cited. The most used metric performance was Accuracy. On average, almost all primary studies presented a result above 90\%, and all primary studies used the Supervisioned type of ML.
\end{abstract}


\section{Introduction}
Population ageing is poised to become one of the most significant social transformations of the 21st century. The number and proportion of people aged 60 and older are increasing. According to the World Health Organization (WHO), the number of people aged 60 and over was 1 billion. This number is increasing to 1.4 billion by 2030, and 2.1 billion by 2050 \citep{who}. This significant change in the global population requires adaptations in all sectors, such as Transportation, Housing, Urban Planning, and Health Care.

Nowadays, the internet is so essential in our lives that it is considered a utility. With this facility of access to the internet, many internet-connected devices are being used in our lives, such as smart TV, home appliances, smartwatch, home security systems, and medical devices. This emergence of connected devices is often referred to as the Internet of Things (IoT). An extension of IoT is the Internet of Medical Things (IoMT), which aims to improve the quality and accessibility of healthcare services using a collection of medical devices and applications that connect to health IT systems. Any internet-connected devices connected to healthcare IT systems are categorized as IoMT devices. The most common are wearable devices such as smartwatches, smart clothing, and other wireless devices (glucose meter, blood pressure monitor, smart oximeter) \citep{gupta2021hierarchical}.

The rapid evolution and implementation of emerging technologies such as Artificial Intelligence (AI) in IoMT are expected to make healthcare services more efficient and thus drive market demand. As a result, according to \citep{iomt_money} the size of the Global IoMT Market is expected to grow from \$60.83 billion in 2019 to \$260.75 billion by 2027, delivering at a Compound Annual Growth Rate (CAGR) of 19.8\% through 2027. In parallel with the IoMT market, the smart textile market is growing \citep{smart_clothe_grow_1,smart_clothe_grow_2}. Smart textiles, also known as e-textiles, are clothes that contain integrated actuators that can connect to a device using Wi-Fi or Bluetooth. There are several fashion categories in the smart textile market, including smart socks \citep{socks_example}, smart pants \citep{pants_exemple}, and smart shirts \citep{shirt_example} where physiological information can be monitored. 

A smart shirt is a type of smart textile that can monitor human activities \citep{id_435,id_54} and detect anomalies \citep{id_15, id_907} through sensors such as accelerometers, gyroscopes, magnetometers, electrocardiograms (ECG), skin temperature, and oxygen saturation (SpO2) \citep{id_910}. With all this data available to analyze, it is possible to identify anomalies using Machine Learning (ML) techniques which can bring significant benefits to users, especially the elderly \citep{vsabic2021healthcare}.

The main objective of this study is to present a Systematic Literature Review (SLR) to verify the state of the art detection of anomalies in smart shirts using machine learning techniques. It is understood as an anomaly is something unusual in the vital signs of the person wearing a smart shirt. For example, it can be an imminent fall, non-standard heartbeat, fever, oxygen saturation far below average, and another anomaly that a specialist can set.

To the best of our knowledge, there are few studies about machine learning for anomaly detection in smart shirts, which was the main reason for this SLR. The primary studies were read thoughtfully and selected based on the methodology of Kitchenham and Charters \citep{keele2007guidelines} concerning the objectives that are: (i) identify machine learning techniques for anomaly detection in the smart shirt; (ii) identify the datasets used to train the ML algorithm; (iii) identify smart shirts or devices for acquiring vital signs; (iv) identify the performance metrics for evaluating the ML model; (v) types of ML being applied. The remainder of this study is divided into five sections: Section 1 provides the main theoretical background. Section 2 describes the methodology used in this SLR. Section 3 lists the results answering this SLR's questions. And finally, Section 5 contains the conclusion.

\section{Theoretical Background}
This section gives an overview of the three main research fields related to this article: Smart Textile, Machine Learning, and Anomaly Detection.

\subsection{Smart Textile}
\begin{figure}[ht]
	\centering
	\includegraphics[width=1\linewidth]{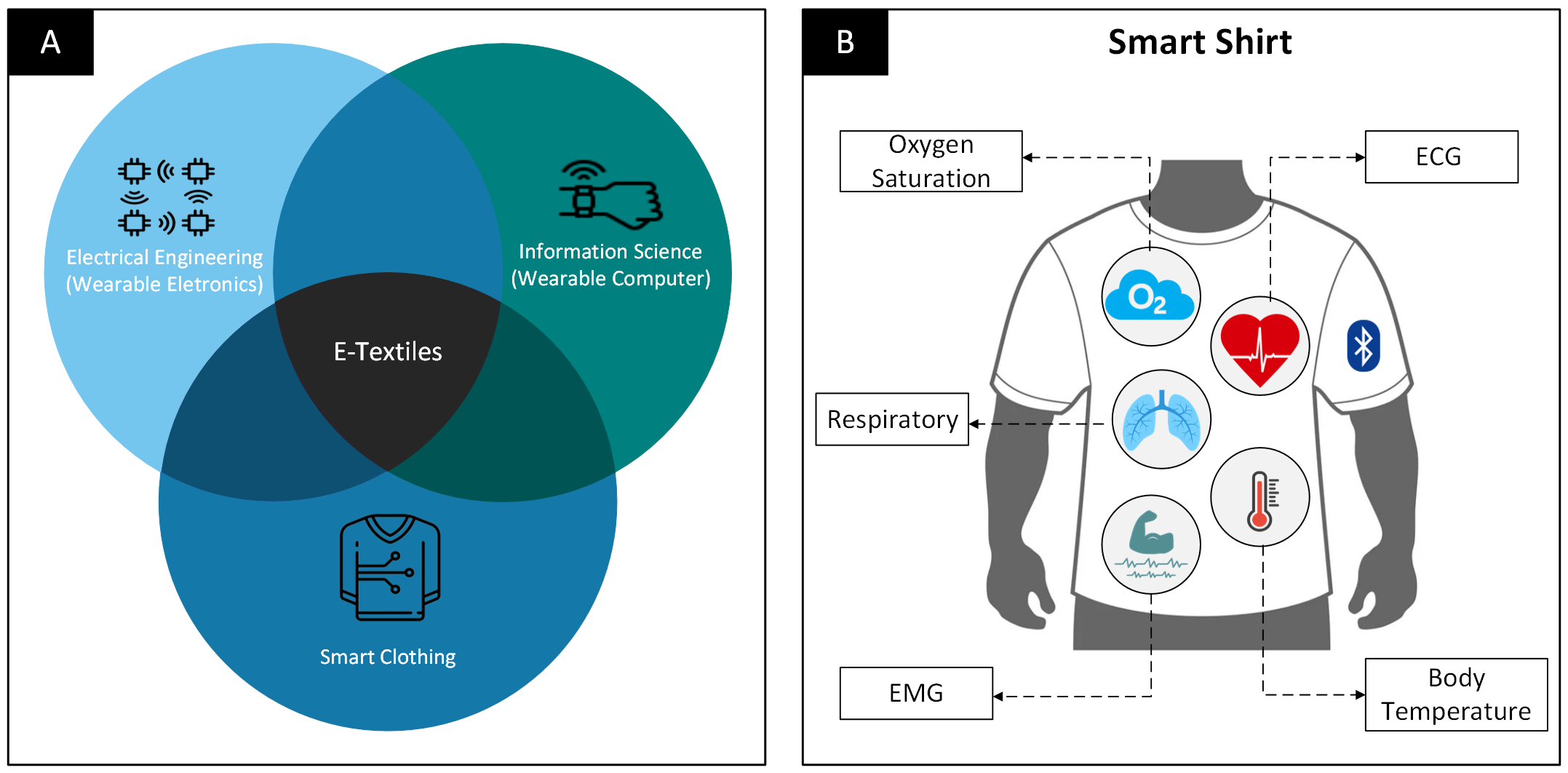}
	\caption{(A) Venn diagram representing the classification of areas in e-textile (adapted from \citep{kan2021future}). (B) Representation of a smart shirt with several sensors.}
	\label{fig:figure_smart_textile}
\end{figure}

In the 1980s, Steve Mann was one of the first researchers to develop smart textile, also called the e-textile, at the Massachusetts Institute of Technology (MIT). This was one of the first attempts to connect hardware to clothing. E-textiles can be classified into three main areas: smart clothing, electrical engineering (wearable electronics), and information science (wearable computers), as shown in Figure \ref{fig:figure_smart_textile}(A) \citep{kan2021future, singha2019recent}.

Smart textile is defined as a product composed of fibres, yarns, filaments, and yarns that integrate electronic components such as sensors, circuits, and communication systems that are powered by some internal or external power supply \citep{khundaqji2020smart, kubicek2020recent}. These communication systems usually use Bluetooth technology, which allows the connectivity of the textile to other smart devices so that it can visualize and analyze data in real-time \citep{khundaqji2020smart}. Smart textile integrate a high level of intelligence and is recognized in three subgroups \citep{kan2021future, kubicek2020recent, stoppa2014wearable}:

\begin{itemize}
	\item \textbf{Passive Smart Textiles}: only able to sense the environment or the user with using textile sensors.
	\item \textbf{Active Smart Textiles}: reactive sensing to stimuli from the environment, which means integrating an actuator function and a sensing device.
	\item \textbf{Very Smart Textile}: able to sense, react and adapt their behavior regarding the ambient conditions.
\end{itemize}

A review study \citep{khundaqji2020smart} showed that smart shirts contain several sensors (e.g. Figure \ref{fig:figure_smart_textile}(b)) for measuring physiological signals such as electrocardiography (ECG), electromyography (EMG), skin temperature, blood oxygen saturation, and respiratory rate. These physiological signals can be used in applications such as healthcare, medicine, transportation and auto use, fashion and entertainment, trackers, and military services \citep{shaveta_bhatia_2020_5529930}.

\subsection{Machine Learning}   
The field of ML has received several formal definitions in the literature.  Samuel defines machine learning as "the field of study that provides computers with the ability to learn, without being explicitly programmed" \citep{samuel1988some}. For Mohri et al.  ML can be broadly defined as computational methods using experience to make predictions, experience (usually in the form of electronic data) referring to past information available for learning \citep{mohri2018foundations}. Mitchell on the other hand provides a small formalism in his definition, "a computer program is said to learn from experience (E), with respect to some class of tasks (T) and performance measure (P), if its performance on tasks in (T), as measured by (P), improves with experience (E)" \citep{Mitchell}. From these definitions, it can summarize that ML is the science of making computers learn and act like humans, improving their learning over time autonomously by feeding them data and information in the form of observations and interactions from the real world.

\begin{figure}[ht]
	\centering
	\includegraphics[width=1\linewidth]{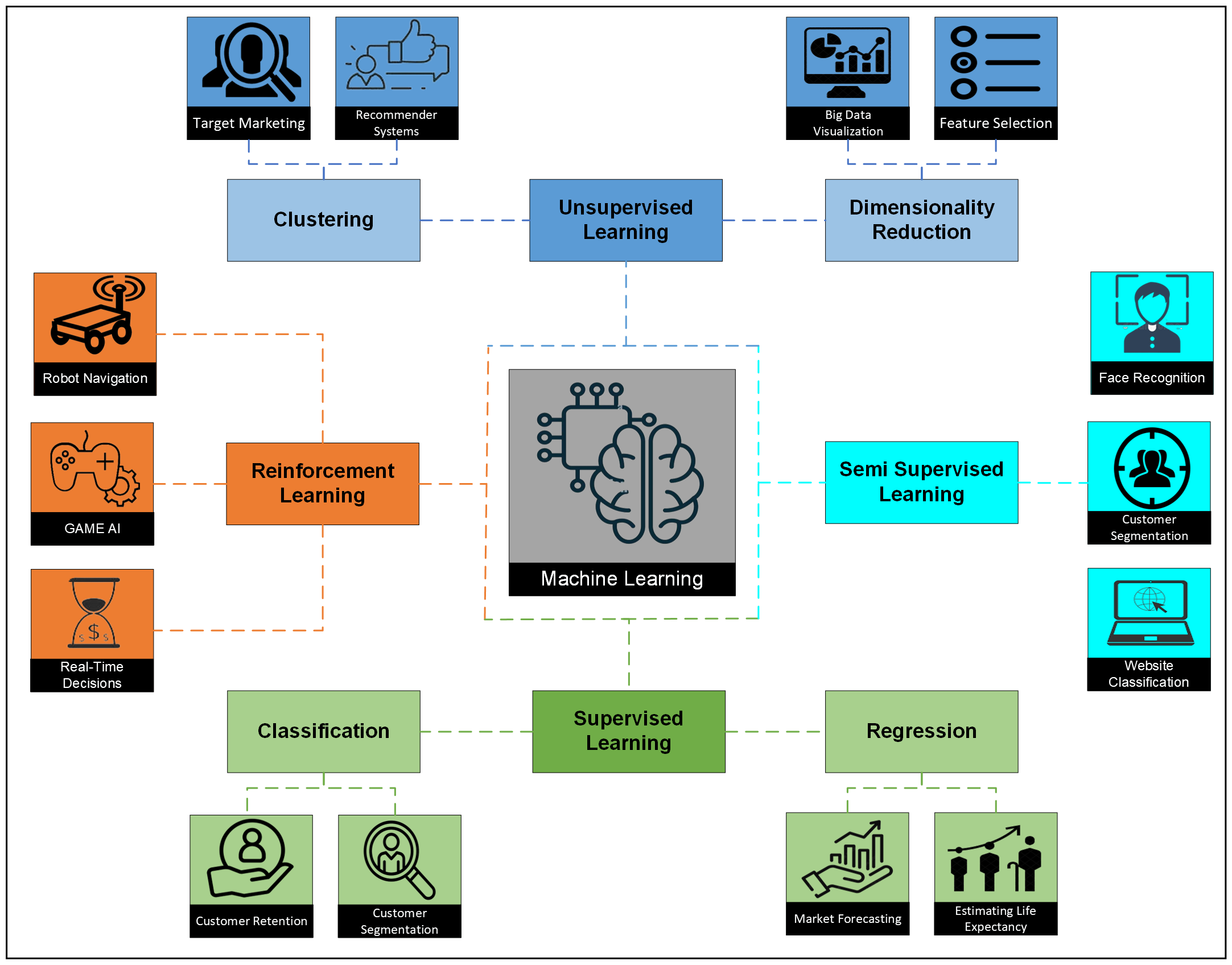}
	\caption{Types of Machine Learning.}
	\label{fig:figure_machine_learning}
\end{figure}

Learning is the process of acquiring knowledge. Unlike us, computers do not learn by reasoning but learn from computer algorithms. Learning is the process of acquiring knowledge. Unlike us, computers do not learn by reasoning but learn from computer algorithms. ML algorithms are organized into taxonomy, based on the desired outcome of the algorithm, as shown in Figure \ref{fig:figure_machine_learning}. Common algorithm type includes \citep{ayodele2010types, russell2016artificial, portugal2018use}:

\begin{itemize}
	\item \textbf{Supervised Learning}: also known as supervised machine learning, is defined by its use of labeled datasets to train algorithms that can classify the data or make predictions. Supervised learning systems, are used to solve regression and classification problems. Examples of some applications: customer retention \citep{customer_retention} and customer segmentation \citep{customer_segmentation} for classification algorithm, and market forecasting \citep{market_forecasting} and estimating life expectancy \citep{estimating_life} for regression algorithm.
	\item \textbf{Unsupervised Learning}: also known as unsupervised machine learning, uses algorithms to analyze and group unlabeled datasets. Examples of some applications: target marketing \citep{target_marketing} and recommender systems \citep{recommender_systems}.
	\item \textbf{Semi-Supervised Learning}: offers a combination of supervised and unsupervised learning where part of the data is partially labeled and the labeled part is used to infer the unlabeled portion. Examples of some applications: face recognition \citep{face_recognition}, customer segmentation \citep{customer_segmenation}, and website classification \citep{website_classification}.
	\item \textbf{Reinforcement Learning}: is a learning model similar to supervised learning, but here the algorithm is not trained using a dataset. The reinforcement learning model learns based on external feedback given either by a thinking entity or by the environment. An analogous approach is teaching dogs to sit or jump. The dog receives a treat (positive feedback) if it performs the action correctly, otherwise it does not receive the treat (negative feedback). Examples of some applications: robot navigation \citep{robot_navigation}, game artificial intelligence \citep{game_ai}, and real time decisions \citep{real_time_decisions}.
\end{itemize}

\subsection{Anomaly Detection}

Anomaly detection is a process to identify unusual patterns that do not conform to expected behaviour, and these unusual patterns are generally referred to as anomalies and outliers \citep{alla2019beginning}. This topic anomaly detection has been used in several fields of study such as data breaches, identity theft, manufacturing, networking, video surveillance, and medicine \citep{alla2019beginning}.

\begin{figure}[ht]
	\centering
	\includegraphics[width=1\linewidth]{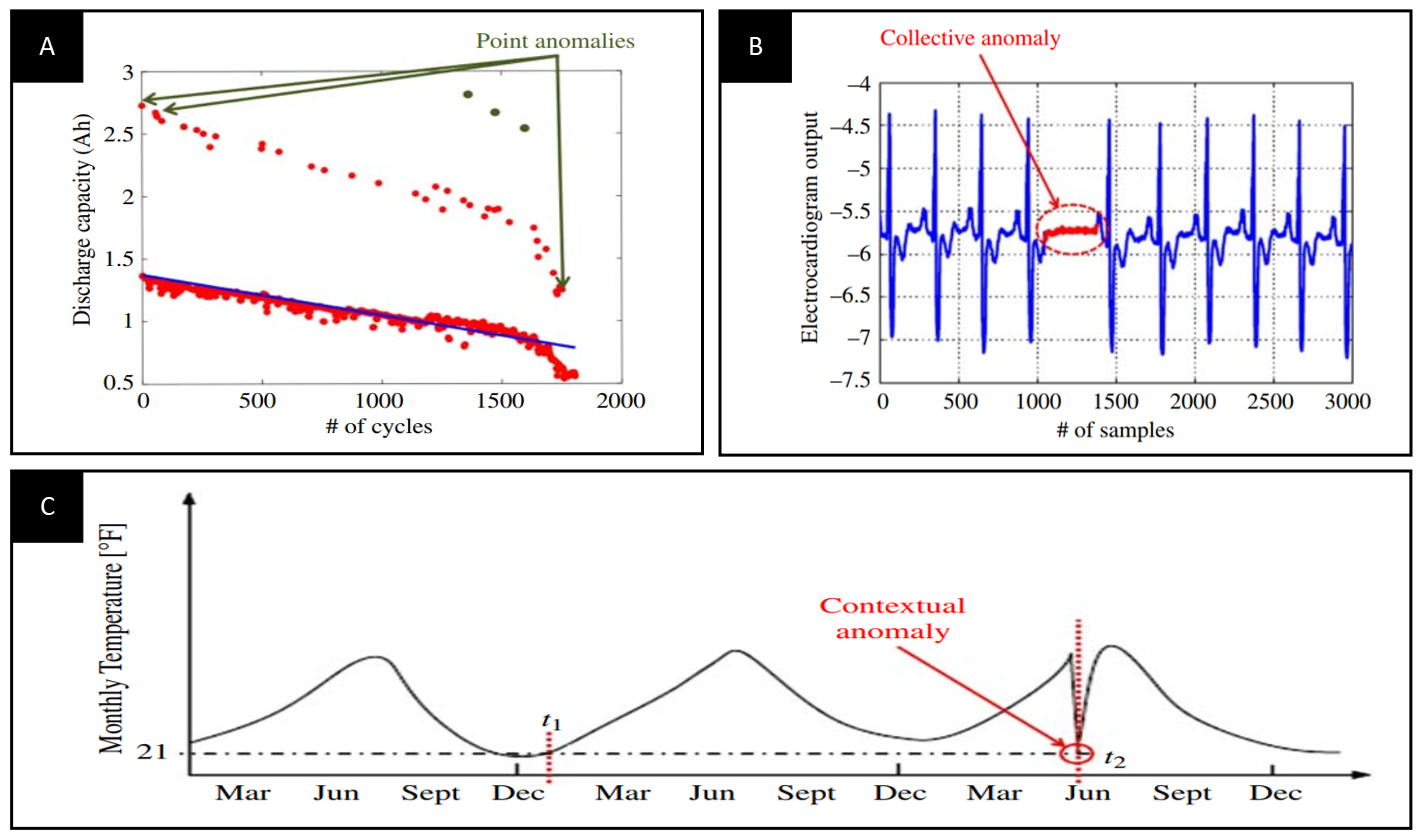}
	\caption{Examples of Categories of Anomalies \citep{pecht2019machine2}.}
	\label{fig:figure_cat_anomaly}
\end{figure}

An important aspect is a proper understanding of the nature of anomalies in the development of anomaly detection methods. Anomalies are classified into three categories \citep{pecht2019machine2, nassif2021machine, alla2019beginning}:

\begin{itemize}
	\item \textbf{Point Anomalies}: A data point-based anomaly is an instance of data that is considered an anomaly for the rest of the data, this type of anomaly is simplest and it is usually the focus of most research on anomaly detection. The Figure \ref{fig:figure_cat_anomaly}(A) is an example of this category where it illustrates the discharge capacity data obtained from a lithium-ion battery and shows the anomaly points. 
	\item \textbf{Contextual Anomalies}: Context-based anomaly is an instance of data that is consider an anomaly if a particular contexto the instance is anomaly, but it is not anomaly in another context. Figure \ref{fig:figure_cat_anomaly}(B) illustrates an example of a temperature time series that shows a monthly temperature for an area. A temperature of 20ºF is considered normal at time \textit{t1} (winter), but a temperature of 20ºF at time \textit{t2} (summer) can be a contextual anomaly.
	\item \textbf{Collective Anomalies}: This category defines that a set of data instances are anomalous in relation to the entire dataset. Figure \ref{fig:figure_cat_anomaly}(C) illustrates an example of an ECG output and the highlighted region is an anomaly set, this is because the human ECG output should not be below for a long time.
\end{itemize}

The use of ML-based anomaly detection is increasingly used, and this technique is used to build a model that distinguishes between normal and abnormal classes \citep{nassif2021machine}. The anomaly techniques can be divided into three categories based on the data function. The three categories are \citep{nassif2021machine, alla2019beginning}:

\begin{itemize}
	\item \textbf{Supervised Anomaly Detection}: requires all instances of the dataset to be labeled "normal" and "anomalous". This technique is basically a type of task that binary classification.
	\item \textbf{Semi-Supervised Anomaly Detection}: requires only instances considered "normal" to be labeled in a dataset. In this technique, the model will only predict normal instances.
	\item \textbf{Unsupervised Anomaly Detection}: does not require instances to be labeled. In these techniques, the model will try to predict which instances are "normal" or "abnormal".
\end{itemize}

\section{Research Method}
\begin{figure}[ht]
	\centering
	\includegraphics[width=1\linewidth]{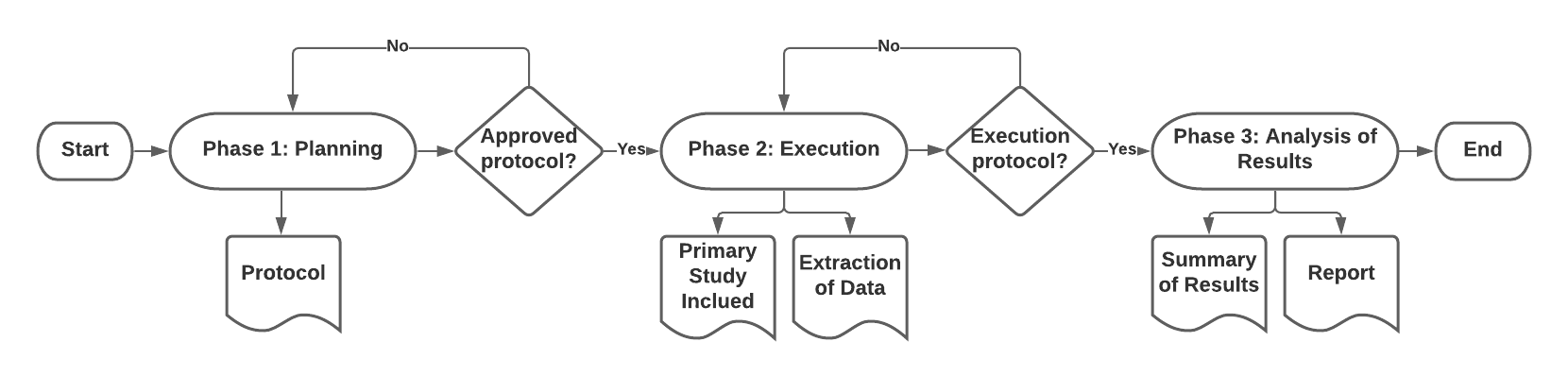}
	\caption{SLR phases adapted \citep{dos2013relationship}.}
	\label{fig:figure_fluxo}
\end{figure}

In this study, a Systematic Literature Review (SLR) was conducted using the methodology of Kitchenham and Charters \citep{keele2007guidelines}. This study was developed considering the three phases: planning, execution, and analysis of results (Figure \ref{fig:figure_fluxo}). In the planning phase, a protocol is defined specifying research questions, keywords, inclusion and exclusion criteria for primary studies, and other topics of interest. In the execution phase, the literature search is conducted following with defined protocol, and this stage that the inclusion and exclusion of primary studies are carried out. Finally, data extraction is done in the results analysis phase, and the results are compared.

\subsection{Research Questions}

This SLR intends to summarize, clarify and examine the ML techniques and implementations applied in anomaly detection in smart shirts from January 2017 through December 2021 inclusive. Table \ref{tab:research_questions} presents four research questions (RQs) are raised for this purpose.

\begin{table}[htpb]
\caption{Research questions.}
\begin{tabular}{p{0.05\textwidth}p{0.45\textwidth}p{0.4\textwidth}}
\toprule
    {\bf RQ\#} & {\bf Research Questions} & {\bf Motivation} \\
\midrule
    $RQ_{1}$ & 
    What anomalies were identified in smart shirt? & 
    Identify which anomaly the primary study tries to detect.  \\
    $RQ_{2}$ & 
    Which ML techniques have been used for anomaly detection in smart shirts? & 
    Aims to identify the ML techniques commonly being used in anomaly detection in smart shirts. \\
    
    $RQ_{3}$ & 
    What kind of empirical validation for anomaly detection in smart shirts is found using the ML techniques found in RQ1? & 
    Assess the empirical evidence obtained. \\
    
    $RQ_{3.1}$ &
    Which datasets are used? & 
    Identify datasets reported to be appropriate.  \\
    
    $RQ_{3.2}$ &
    Which devices are used for data acquisition? & 
    Identify devices being used.  \\
    
    $RQ_{3.3}$ &
    Which performance measures are used? & 
    Identify the performance of the ML techniques.  \\
    
    $RQ_{4}$ &
    What is the overall performance of the ML techniques for anomaly detection in smart shirts? & 
    Analyze the results from the performance metric.  \\
    
    $RQ_{5}$ &
    What types of ML algorithms are being applied in anomaly detection in smart shirts? & 
    Specify which ML types have been applied. \\
\bottomrule
\label{tab:research_questions}
\end{tabular}
\end{table}

\subsection{Search Strategy}
The search terms and synonyms were formed according to related studies using machine learning, anomaly detection, and smart textile. All these terms and synonyms were included in the Search Query (SQ), which is presented as follows:

\textit{query\_01 = ("Smart Shirt" \textbf{OR} "Smart Clothing" \textbf{OR} "Smart Clothes" \textbf{OR} "Smart Cloth" \textbf{OR} "Smart Textiles" \textbf{OR} "Smart Textile" \textbf{OR} "e-textile" \textbf{OR} "Clothing Tech" \textbf{OR} "Wearable Smart Textile" \textbf{OR} "e-clothes" \textbf{OR} "clothing tech")}

\textit{query\_02 = ("Anomaly Detection" \textbf{OR} "Anomalous Detection" \textbf{OR} "Abnormal Detection" \textbf{OR} "Outlier Detection" \textbf{OR} "Machine Learning" \textbf{OR} "Artificial Intelligence" \textbf{OR} "Expert System" \textbf{OR} "Intelligence System" \textbf{OR} ”Supervised Anomaly Detection” \textbf{OR} ”Semi-Supervised Anomaly Detection” \textbf{OR} ”Unsupervised Anomaly Detection”)}

\begin{center} \textit{SQ = query\_01 \textbf{AND} query\_02}\end{center}

After identifying the search terms, important digital portals were selected. The following six electronic databases were used to search the primary studies:

\begin{itemize}[noitemsep,topsep=0pt]
	\item \textbf{ACM Digital Library}: https://https://dl.acm.org/
	\item \textbf{Scopus}: http://www.scopus.com
	\item \textbf{Google Scholar}: https://scholar.google.com/
	\item \textbf{PubMed}: https://pubmed.ncbi.nlm.nih.gov/
	\item \textbf{Web of Science}: https://www.webofknowledge.com/
	\item \textbf{ScienceDirect}: https://www.sciencedirect.com/
\end{itemize}

\subsection{Study Selection}

\begin{figure}[ht]
	\centering
	\includegraphics[width=1\linewidth]{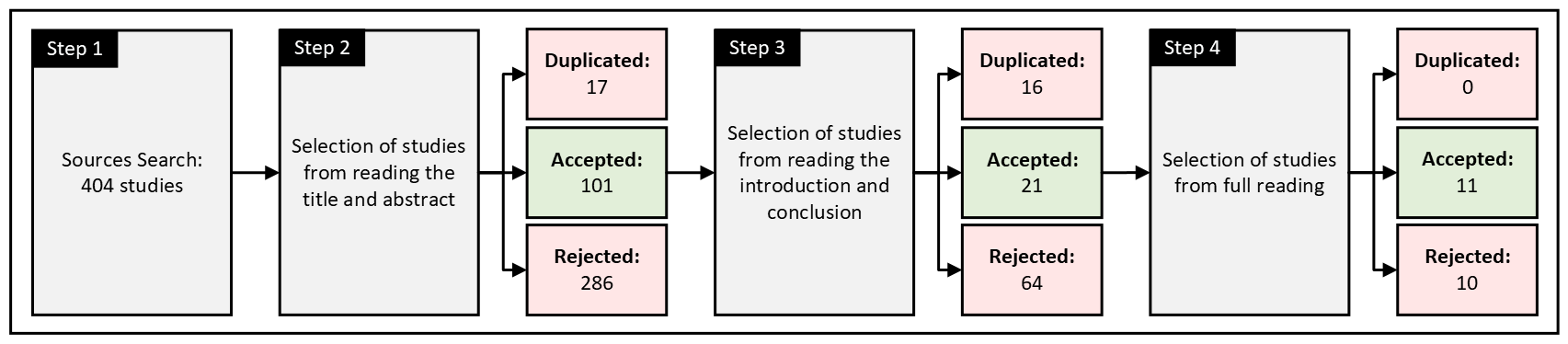}
	\caption{Selection Process of Primary Studies \citep{dos2013relationship}.}
	\label{fig:slr_process}
\end{figure}

The whole process of filtering and searching the primary studies is illustrated in Figure \ref{fig:slr_process}. Searches were carried out in the digital databases using the SQ described in Section 3.2, where studies in the determined interval ($IC_{1}$) were selected. This first step has collected 404 studies, of which 121 came from ScienceDirect, 88 from ACM Digital Library, 76 from Scopus,  56 from Google Scholar, 49 from Web of Science, and 14 from PubMed.

The inclusion (IC) and exclusion (EC) criteria summarized in Table \ref{tab:exclusion_inclusion} were applied to select the studies in steps two, three, and four. In the second step, the title and abstract of all selected studies were read, where there were 17 duplicate studies, 286 rejected studies, and 101 accepted studies. In the third step, the introduction and conclusion of the studies accepted in the previous step were read, where there were 16 duplicate studies, 64 rejected studies, and 21 accepted studies. In the fourth step, the studies accepted in the previous step were read in their entirety, where there were no duplicated studies, 10 rejected studies, and finally 11 accepted studies.

\begin{table}[htpb]
\caption{Inclusion and exclusion criteria.}
\begin{tabular}{p{0.1\textwidth}p{0.8\textwidth}}
\toprule
    {\bf Inclusion} & {\bf criteria} \\ 
\midrule
    $IC_{1}$ & 
    Studies published in the range 2017-2021. \\ 
    $IC_{2}$ &
    Studies written in English. \\
    $IC_{3}$ &
    Studies use ML techniques for anomaly detect in smart textile. \\
    $IC_{4}$ &
    Studies primary. \\
    
\toprule
    {\bf Exclusion} & {\bf criteria} \\
\midrule    
    $EC_{1}$ &
    Studies with no clear publication information. \\
    $EC_{2}$ &
    Studies that do not use smart textile. \\
    $EC_{3}$ &
    Studies that do not use ML techniques for anomaly detection. \\
    $EC_{4}$ &
    Studies that do not use techniques ML in smart textile. \\
    $EC_{5}$ &
    Studies that are not written in English. \\
    $EC_{6}$ &
    Studies that do not provide the full paper. \\
    $EC_{7}$ &
    Studies that do not present the abstract. \\ 
    $EC_{8}$ &
    Studies in which year of publication is before 2017. \\
    $EC_{9}$ &
    Books, letters, notes, and patents are not included in the review. \\
\bottomrule
\label{tab:exclusion_inclusion}
\end{tabular}
\end{table}

\subsection{Quality Assessment Rules}

The Quality Assessment Rules (QARs) are fundamental to ensure and assess the quality of the primary studies and are performed in the final step to define which studies will be added for review. Inspired by the study \citep{nassif2021machine}, 10 QARs were assigned (Table \ref{tab:QRAs}) where each QAR can be assigned a score from 1 to 10. Following the same method as in the study \citep{nassif2021machine}, studies will only be accepted if they obtain a score of 5 or more, otherwise, the study is rejected.

\begin{table}[htpb]
\caption{Quality assessment questions.}
\begin{tabular}{p{0.1\textwidth}p{0.8\textwidth}}
\toprule
    {\bf Q\#} & {\bf Quality questions} \\ 
\midrule
    $QAR_{1}$ & 
    Are the study objectives clearly recognized? \\
    $QAR_{2}$ &
    Are the anomaly detection techniques well defined and deliberated? \\
    $QAR_{3}$ &
    Is the specific application of anomaly detection clearly defined? \\
    $QAR_{4}$ &
    Does the paper cover practical experiments using the proposed technique? \\
    $QAR_{5}$ &
    Are the experiments well designed and justifiable? \\
    $QAR_{6}$ &
    Are the experiments applied on sufficient datasets? \\ 
    $QAR_{7}$ &
    Are estimation accuracy criteria reported? \\
    $QAR_{8}$ &
    Is the proposed estimation method compared with other methods? \\
    $QAR_{9}$ &
    Are the techniques of analyzing the outcomes suitable? \\
    $QAR_{10}$ &
     Overall, does the study enrich the academic community or industry? \\
\bottomrule
\label{tab:QRAs}
\end{tabular}
\end{table}

\subsection{Data Extraction and Data Synthesis}
The StArt tool (State of the Art through Systematic Review) \citep{fabbri2016improvements} was used to support the SLR process, mainly in data extraction and data synthesis.  During the data extraction step, within the StArt tool, it was possible to summarize the primary studies such as author name, study title, publication details, devices used in the smart shirt, dataset, evaluation metrics, and machine learning and anomaly techniques.

In summary, the goal in synthesizing data is to accumulate and combine facts and figures from the selected primary studies to reformulate answers and address the research questions \citep{wen2012systematic}. To answer these questions, visualization techniques such as line graphs, bar graphs, and box plots were used. Tables were also used to summarize and present the results of this SLR.

\section{Result and Discussions}
This section presents the results obtained from selected primary studies. First, an overview of the selected works is presented and then the results for each research question in this SLR.

\subsection{Description of Primary Studies}
After following all the steps of the SLR, Table \ref{tab:primary_study} shows all the accepted primary studies. In total, 11 were considered for meta-analysis. Various graphs and tables are normally used for meta-analysis, but as in this study there were few primary studies accepted, it is possible to see in Table \ref{tab:primary_study} information about the primary studies such as the assignment of an ID to each study, title, year of publication, type, and its respective reference.

\begin{table}[htpb]
\caption{Selected primary studies.}
\begin{tabular}{p{0.05\textwidth}p{0.65\textwidth}p{0.05\textwidth}p{0.05\textwidth}p{0.05\textwidth}}
\toprule
    {\bf ID} & {\bf Title}  & {\bf Year}  & {\bf Type} & {\bf Refs.}\\ 
\midrule
    A1 & "SR-ScatNet Algorithm for On-device ECG Time Series Anomaly Detection" & 2021 & Conf. & \citep{id_15} \\ \\
    A2 & "Physical Activity Recognition Based on a Parallel Approach for an Ensemble of  Machine Learning and Deep Learning Classifiers." & 2021 & Jour. & \citep{id_435} \\ \\
    A3 & "Sensor Shirt as Universal Platform for Real-Time Monitoring of Posture and Movements for Occupational Health and Ergonomics" & 2021 & Jour. & \citep{a3} \\ \\ 
    A4 & "A New Smart-Fabric based Body Area Sensor Network for Work Risk Assessment" & 2020 & Conf. & \citep{a4} \\ \\
    A5 & "Violent Activity Recognition by E-textile Sensors based on Machine Learning Methods" & 2020 & Jour. & \citep{a5} \\ \\ 
    A6 & "Smart Healthcare Framework for Ambient Assisted Living using IoMT and Big Data Analytics Techniques" & 2019 & Jour. & \citep{a6} \\ \\ 
    A7 & "Application of Hierarchical Temporal Memory to Anomaly Detection of Vital Signs for Ambient Assisted Living" & 2019 & Thes. & \citep{a7} \\ \\
    A8 & "An Investigation of Different Machine Learning Approaches for Epileptic Seizure Detection" & 2019 & Conf. & \citep{a8} \\ \\ 
    A9 & "Physical Activity Classification using a Smart Textile" & 2018 & Conf. & \citep{id_54}  \\ \\
    A10 & "Fall Detection System for Elderly People using IoT and Big Data" & 2018 & Jour. & \citep{a10} \\ \\
    A11 & "Context Aware Adaptable approach for Fall Detection bases on Smart Textile" & 2017 & Jour. & \citep{a11} \\
\bottomrule
\label{tab:primary_study}
\end{tabular}
\end{table}

A1: Feng \textit{et al.} \citep{id_15} propose an architecture called SR-ScatNet to detect abnormal ECG signals in smart shirts. The authors used the MIT-BIH Arrhythmia Database to validate the proposed architecture where it obtained 98\% accuracy for abnormal signals.

A2: The focus of the study by Abid \textit{et al} \citep{id_435} is to investigate Human Activity Recognition (HAR) using a smart shirt. The study detects 10 types of activities and one of them is fall detection which is an important anomaly in a smart shirt. The experiment involved 44 people and they used 3 machine learning techniques for HAR.

A3: The study by Petz \textit{et al} \citep{a3} describe the development of a smart shirt for recording upper body movement and shows that it is possible to perform posture recognition. Results show the potential of the smart shirt for applications in the field of occupational safety.

A4: Lanata \textit{et al} \citep{a4} propose a sensor suite with a smart shirt for physiological work risk levels. The combination of the wireless sensor and smart fabric allows for worker risk assessment. This study focuses on the description of the sensors and proposes a k-nearest neighbors (KNN) algorithm for risk assessment and detection.

A5: Randhawa \textit{et al} \citep{a5} study aims to recognize if a person is being physically attacked through sensors installed in a smart shirt. The study analyzed the best sensor positions and applied several ML techniques to recognize physical aggressions where these aggressions can be considered anomalies.

A6: Syed \textit{et al} \citep{a6} propose a healthcare framework to monitor the physical activities of the elderly using IoMT and ML algorithms. In total it is proposed to recognize 12 types of activities, and one of the activities to recognize is the elderly's fall, which can be considered an anomaly. 

A7: The study is a Ph.D. thesis by Bastaki \citep{a7} where it is presents the development of a framework for vital sign anomaly detection for an Ambient Assisted Living (AAL) health monitoring scenario. This study created its own dataset and used three ML techniques.

A8: Resque \textit{et al} \citep{a8} investigate the performance of five ML algorithms in terms of accuracy for Epileptic Seizure Detection. The study concluded that the ML techniques had good accuracy for identifying epileptic seizures from EEG that can also be considered an anomaly in smart shirts.

A9: The aim of the study by Cherif \textit{et al} \citep{id_54} is to develop a human activity classification system using a smart shirt. The study used ML techniques. One of the activities is falling which can be considered an anomaly.

A10: Yacchirema \textit{et al} \citep{a10} built a framework for fall detection in the elderly. If the system detects a fall, an alert is issued and the system automatically sends it to the groups responsible for providing care to the elderly. The result, the study found high success rates in fall detection in terms of accuracy, precision, and gain.

A11: The study Mezghani \textit{et al} \citep{a11} presents a novel fall detection system using smart textile and ML techniques. The system classifies falls based on their respective orientations among 11 classes. The results showed that the system is 98\% accurate for fall detection.

\subsection{RQ1: What anomalies were identified in smart shirt?}

The anomalies of the primary studies were identified and are shown in Table \ref{tab:tab_anomalies}. The most frequent anomaly is the fall present in five studies. A fall is an unintentional displacement of the body to a lower level than the initial position, with the inability to correct it promptly caused by multifactorial circumstances. Fall-related injuries may be fatal or non-fatal.

Primary study A1 addresses the anomaly related to cardiac arrhythmia. Cardiac arrhythmia is characterized by a lack of rhythm in a heartbeat. Tachycardias is when the rhythm is fast, and bradycardias are when the rhythm is too slow, both of which can worsen and lead to heart complications. The study A3 addresses posture recognition where it can be considered an anomaly because postural diseases are caused by poor posture \citep{a3}. The A5 study presents a smart shirt to detect violent attacks or aggression. The study A7 uses a set of vital signs for monitoring and detecting anomalies in patients.The detection of epilepsy is addressed in study A8 is also considered an anomaly that can issue an alert if the patient is in a crisis. Moreover, finally, the A4 study is a development of a smart shirt to detect risk at work, and the anomaly it wanted to detect was not well defined.

\begin{table}[htpb]
\centering
\caption{Anomalies identified in primary studies.}
\begin{tabular}{p{0.2\textwidth}p{0.15\textwidth}p{0.2\textwidth}}
\toprule
    {\bf Anomaly} & {\bf \# of Studies} & {\bf ID}\\ 
\midrule
    Fall & 5 & A2, A6, A9, A10, A11 \\
    Cardiac Arrhythmia & 1 & A1 \\
    Posture & 1 & A3 \\
    Violent Attack & 1 & A5 \\
    Vital Signs & 1 & A7 \\
    Epilepsy & 1 & A8 \\
    Undefined & 1 & A4 \\ 
\bottomrule
\label{tab:tab_anomalies}
\end{tabular}
\end{table}

\subsection{RQ{2}: Which ML techniques have been used for anomaly detection in smart shirts?}
This section provides details of the ML techniques used in the primary studies selected in this SLR. With the analysis of the selected studies, the following ML techniques were identified shown in Table \ref{tab:method_ML}. 

\begin{table}[htpb]
\caption{Distribution os studies across ML techniques .}
\begin{tabular}{p{0.35\textwidth}p{0.15\textwidth}p{0.15\textwidth}p{0.25\textwidth}}
\toprule
    {\bf Method} & {\bf \# of Studies} & {\bf Percent}  & {\bf ID}\\ 
\midrule
    Support Vector Machines (SVM) & 7 & 63.63\% & A1, A2, A5, A7, A8, A9, A11 \\
    K-Nearest Neighbors (KNN) & 6 & 54.54\% & A1, A4, A5, A7, A8, A9 \\
    Naive Bayes (NB) & 3 & 27.27\% & A5, A6, A8 \\
    Decision Trees (DT) & 2 & 18.18\% & A5, A10  \\
    Linear Discriminant Analysis (LDA) & 2 & 18.18\% & A1, A5 \\
    Convolutional Neural Network (CNN) & 2 & 18.18\% & A1, A2 \\
    Long Short-Term Memory (LSTM) & 1 & 09.08\% & A3 \\
    Principal Component Analysis (PCA) & 1 & 09.08\% & A5 \\
    Hierarchical Temporal Memory (HTM) & 1 & 09.08\% & A7 \\
    Random Forest (RF) & 1 & 09.08\% & A8 \\
    Neural Network (NN) & 1 & 09.08\% & A8 \\
\bottomrule
\label{tab:method_ML}
\end{tabular}
\end{table}

Table \ref{tab:method_ML} presents the number and percentage of studies in relation to ML techniques. The analyzed ML techniques frequently used were SVM with seven or 63\% of the studies and KNN with six or 54\% of the selected studies. The NB, DT, LDA, and CNN techniques were used in three (27\%), two (18\%), two (18\%), two (18\%) studies, respectively. Moreover, the LSTM, PCA, HTM, RF, and NN techniques were less used.

\subsection{RQ3: What kind of empirical validation for anomaly detection in smart shirts is found using the ML techniques found in RQ1?}
This section aims to identify the datasets, the devices for data acquisition, and the performance measures used in the selected primary studies.

\subsubsection{RQ3.1: Which datasets are used?}
Data is an essential component of any ML technique. The dataset can be used to train an ML algorithm that aims to find predictable patterns from the entire dataset. In total, 11 datasets were identified, which are shown in the Table \ref{tab:table_table_dataset_ml}. Each primary study used only a dataset, where five studies created and used their own dataset, and six studies used public and private datasets.

\begin{table}[htpb]
\caption{Description of the datasets found in this SLR.}
\begin{tabular}{p{0.03\textwidth}p{0.1\textwidth}p{0.8\textwidth}}
\toprule
    {\bf ID} & {\bf Dataset} & {\bf Description}\\ 
\midrule
    A1 & MIT-BIH Arrhythmia \citep{dataset_a1} &  Contains 48 half-hour excerpts of two-channel ambulatory ECG recordings, obtained from 47 subjects. It provides around thirty-four different ECG samples. \\ \\
    
    A2 & Own Dataset &  Dataset used in this study is acquired by the research team. In the experiments, 44 participants wearing the Hexoskin shirt, performed 10 activities. \\ \\
    
    A3 & Own Dataset & The study does not describe the dataset. The dataset contains three positions 'leaning back', 'straight', and 'leaning forward' were defined and held for 5 seconds or 50 data samples. \\ \\
    
    A4 & Own Dataset & - \\ \\
    
    A5 & Human Motion & Composed of 5K-10K samples distributed in stationary, walking, brisk walking, twisting, and aggressive motion of five classes. \\ \\
    
    A6 & MHEALTH \citep{dataset_a6} & Comprises body motion and vital signs recordings for ten volunteers in several physical activities. Sensors (2-lead ECG) are placed on different parts of the body to measure acceleration, rate of turn, and magnetic field orientation.   \\ \\
    
    A7 & MIMIC-III & It contains data related to health-related of more than forty thousand patients who were in critical care. Data includes demographics, vital signs, laboratory results, medications, etc. \\ \\
    
    A8 & EEG UCI & Contains five classes ( seizure activity, tumor area, healthy brain area, patients had their eyes closed, patients had their eyes open) that used EEG signals. \\ \\
    
    A9 & Own Dataset & Contains thirteen volunteers, each participant used a smart shirt to build the dataset. \\ \\
    
    A10 & SisFall \citep{dataset_10} & It is a dataset of falls and activities of daily living that was acquired from a device composed of two types of accelerometers and a gyroscope. \\ \\ 
    
    A11 & Own Dataset & It contains thirteen volunteers where each volunteer wore a smart shirt and performed 11 activities. It was monitored: acceleration data, cardiac activity, and respiratory activity were monitored. \\
\bottomrule
\label{tab:table_table_dataset_ml}
\end{tabular}
\end{table}

\subsubsection{RQ3.2: Which Devices are used for data acquisition?}

Sensors capture data from the world around us, and these sensors are integrated into devices so that data can be displayed, analyzed and stored. Using a smart shirt, it is possible to capture vital signs and measure the body's inertia. Table \ref{tab:sensors} shows the type of smart textile and the sensors used from the selected primary studies.

\begin{table}[htpb]
\caption{Description of the datasets found in this SLR.}
\begin{tabular}{p{0.02\textwidth}p{0.18\textwidth}p{0.73\textwidth}}
\toprule
    {\bf ID} & {\bf Type} & {\bf Sensor Data}\\ 
\midrule
    A1 & Smart Shirt Prototype & ECG \\
    A2 & Hexoskin Smart Shirt & Accelerometer \\
    A3 & Smart Shirt Prototype & Accelerometer, Gyroscope, and Magnetometer \\
    A4 & Smart Shirt Prototype & ECG, Respiration Sensor, Accelerometer, Gyroscope, Magnetometer, and Sweat Sensor \\
    A5 & Smart Shirt Prototype & Pressure Sensor, Stretch Sensor, and Accelerometer \\
    A6 & Wearable & ECG, Accelerometer, Gyroscope, and Magnetometer \\
    A7 & Wearable & Heart Rate, Arterial Blood Pressure, and Respiratory Rate \\
    A8 & Wearable & ECG \\
    A9 & Hexoskin Smart Shirt & Accelaration, Cardic Activity, and Respiratory Activity \\
    A10 & Wearable & Accelerometer, Gyroscope, and Magnetometer \\
    A11 & Hexoskin Smart Shirt & Accelaration \\
\bottomrule
\label{tab:sensors}
\end{tabular}
\end{table}

In this SLR, seven smart shirts were identified and only one (Hexoskin) is being sold commercially, the rest are proposals or prototypes. The remaining were classified as wearables, but with one exception, it is wearables that are located in the human's chest. 

Most of the primary studies use an accelerometer sensor in which it is possible to measure the person's acceleration and an ECG sensor in which it is possible to measure electrical impulses from the heart using a smart shirt. The pressure and sweat sensors were the least used sensors, as they were used in only one study.

The accelerometer sensor is the most used because most of the primary studies are related to the recognition of human activity where one of the activities is fall. However, some selected primary studies aim to solve specific problems. It is common to choose specific sensors, so the less identified sensors are used for particular problems, for example, in study A4.

\subsubsection{RQ3.3: Which performance metrics are used?}
When implementing an ML technique, the developer must ask himself how good the technique is for anomaly detection. All ML techniques need a metric to evaluate performance. Thus, in total, seven metrics were identified in the primary studies, which are shown in Table \ref{tab:performance_metric}.

\begin{table}[htpb]
\caption{Definition of performance metrics related to selected primary studies.}
\begin{tabular}{p{0.2\textwidth}p{0.6\textwidth}p{0.13\textwidth}}
\toprule
    {\bf Performance Metric} & {\bf Definition} & {\bf ID}\\ 
\midrule
    Confusion Matrix & The confusion matrix is not exactly a performance metric, but it is a base that other metrics use. It is a table that indicates the errors and successes of the ML model where each cell represents an evaluation factor that are True Positive (TP), True Negative (TN), False Positive (FP), and False Negative (FN). & A1, A2, A3, A5, A6, A10 \\ \\
    
    Accuracy & It is the proportion of the total number of correct predictions divided by the total number of predictions. & A1, A3, A5, A6, A8, A9, A10, A11  \\ \\
    
    Precision & It is a metric that measures the amount of true positives (TP) divided by the sum of all positive values (TP+FP) & A2, A6, A7, A10 \\ \\
    
    Recall & Evaluate the ability of the method to successfully detect results classified as positive (True Positive Rate). & A2, A6, A7, A10 \\ \\
    
    F1-Score & It is a harmonic mean calculated based on precision and recall. & A2, A7 \\ \\
    
    Specificity & It evaluates the ability of the method to detect negative results (True Negative Rate). & A11 \\
\bottomrule
\label{tab:performance_metric}
\end{tabular}
\end{table}

Some studies contained the confusion matrix to illustrate the performance of the ML technique. The most used performance metric is Accuracy, followed is closely followed by the Precision and Recall measures. Specificity and F1-Score are other commonly used metrics. The Kappa metric is less used cited in a primary study. The primary study A4 did not mention any metric performance, but as it uses KNN as ML techniques, it could possibly use the metrics shown in Table \ref{tab:performance_metric}.

\subsection{RQ4: What is the overall performance of the ML techniques for anomaly detection in smart shirts?}

Table \ref{tab:result_metric}  shows all results, types, ML models and performance metrics. Most techniques are classic ML algorithms that do not require significant computational power. Almost all algorithms present satisfactory results in their respective studies.

Primary study A4 did not mention the result because the main objective was to show the development of the smart shirt. After that, according to the authors, they will do a study focusing on machine learning techniques. The primary study A3 did not present an exact value, and it only showed that according to the tests, it obtained an accuracy greater than 80\%.

\begin{table}[htpb]
\caption{Performance metrics and respective results of the selected primary studies.}
\begin{tabular}{p{0.03\textwidth}p{0.1\textwidth}p{0.3\textwidth}p{0.3\textwidth}p{0.1\textwidth}}
\toprule
    {\bf ID} & {\bf Type} & {\bf ML Model} & {\bf Performance Metrics} & {\bf Results}\\ 
\midrule
    A1 & Supervised & CNN + SVM & Accuracy & 98.30\% \\ \\
    
    A2 & Supervised & SVM & Precision & 94\% \\
    & & & Recall & 92\% \\
    & & & F1-Score & 92\% \\
    & & LDA + KNN & Precision & 93\% \\
    & & & Recall & 83\% \\
    & & & F1-Score & 87\% \\
    & & CNN & Precision & 87\% \\
    & & & Recall & 98\% \\
    & & & F1-Score & 97\% \\ \\
    
    A3 & Supervised & LSTM & Accuracy & >80\% \\ \\
    
    A4 & Supervised & KNN & - & - \\ \\
    
    A5 & Supervised & KNN & Accuracy & 93.80\% \\
    & & NB & Accuracy & 96.90\% \\
    & & DT & Accuracy & 93\% \\
    & & SVM & Accuracy & 97.60\% \\ \\
    
    A6 & Supervised & NB & Accuracy & 97.10\% \\
    & & & Precision & 96.38\% \\
    & & & Recall & 97\% \\ \\
    
    A7 & Supervised & HTM & F1-Score & 83\% \\
    & & & Recall & 87\% \\
    & & & Precision & 70\% \\
    & & KNN & F1-Score & 86\% \\
    & & & Recall & 90\% \\
    & & & Pecision & 82\% \\
    & & SVM & F1-Score & 54\% \\
    & & & Recall & 27\% \\
    & & & Precision & 90\% \\ \\
    
    A8 & Supervised & SVM & Accuracy & 97.31\% \\
    & & RF & Accuracy & 97.08\% \\
    & & NB & Accuracy & 95.98\% \\
    & & NN & Accuracy & 93.53\% \\
    & & KNN & Accuracy & 90.01\% \\ \\
    
    A9 & Supervised & KNN & Accuracy & 96.37\% \\
    & & SVM & Accuracy & 95.40\% \\ \\
    
    A10 & Supervised & DT & Accuracy & 91.67\% \\
    & & & Precision & 93.75\% \\
    & & & Recall & 91.67\% \\ \\
    
    A11 & Supervised & SVM & Accuracy & 98\% \\
    & & & Sensitivity & 97.6\% \\
    & & & Specificity & 98.5\% \\
\bottomrule
\label{tab:result_metric}
\end{tabular}
\end{table}

\subsection{RQ5: What types of ML algorithms are being applied in anomaly detection in smart shirts?}

The purpose of this question is to identify the type of ML and anomaly detection for each selected primary study. Table \ref{tab:result_metric} shows that all selected primary studies are of the supervised type. Being of the supervised type is that all studies have classification or multiclassification problems.

\section{Conclusion}

This paper presents a systematic literature review in order to identify machine learning techniques for anomaly detection in smart shirts. First, the theoretical background of smart textile, machine learning and anomaly detection was described. Second, the adopted research method was described following a systematic series of steps analyzing the quality of the primary studies. In total, eleven primary studies (2017-2021) were selected for analysis. And finally, the results, which are the answers to the research questions of the systematic review, are presented.

From the selected primary studies, six different anomalies were identified, which are: fall, cardiac arrhythmia, posture, violent attack, vital signs and ellipsis, with the fall anomaly being the most frequent being cited in five primary studies. Therefore, it is possible to define a set of anomalies to be detected in smart shirts.

About the machine learning techniques for anomaly detection in smart shirts, eleven different techniques were identified: Support Vector Machines, K-Nearest Neighbors, Naive Bayes, Decision Trees, Linear Discriminant Analysis, Convolutional Neural Network, Long Short-Term Memory, Principal Component Analysis, Hierarchical Temporal Memory, Random Forest, and Neural Network, with the Support Vector Machines technique being the most frequent in seven primary studies.

For the machine learning algorithm training, five primary studies created their dataset and six primary studies used public or private datasets, with each study using a dataset. In general, no dataset was cited more than once.

Evaluating the performance of the machine learning algorithm is a fundamental process. Five different metrics were identified. The accuracy of the performance metric was the most cited metric in eight primary studies. Overall, almost all primary studies showed high (>90\%) and good accuracy values on their respective problems. It was also observed that the primary studies used the confusion matrix to validate and show the performance of the machine learning algorithm.

Finally, seven primary studies used or built a smart shirt regarding data acquisition devices, and four studies used a wearable simulating a smart shirt. A smart shirt can contain one or more sensors, and the accelerometer sensor was the most cited since most studies involve fall detection.

This study aims to result in a second study that can contribute some direction in research related to the topic. Furthermore, it is hoped with future work to implement an intelligent system of anomaly detection in smart shirts based on the results of this analysis.

\newpage

\bibliographystyle{unsrtnat}

\end{document}